\documentclass[letterpaper, 10 pt, conference]{ieeeconf}  

\IEEEoverridecommandlockouts                              

\usepackage{siunitx}
\usepackage{graphicx}
\usepackage{float}
\usepackage{circledsteps}
\usepackage{booktabs}
\usepackage{tikz}
\usepackage{amsmath}
\usepackage{subfig}
\usepackage{gensymb}
\usepackage{cite}

\DeclareSIUnit\revolution{rev}

\definecolor{UStuttGrey}{RGB}{62,68,76}
\newdimen\satlevel
\newdimen\satdiameter
\newcommand{\rating}[1]{%
	\satdiameter=2.5ex\relax
	\ifcase#1\relax
	\satlevel=0pt\relax
	\or
	\satlevel=0.22\satdiameter
	\or
	\satlevel=0.3\satdiameter
	\or
	\satlevel=0.5\satdiameter
	\fi
	\tikz[baseline=-0.3\satdiameter]{%
		\fill[UStuttGrey] (0,0) circle (\satlevel);
		\path [draw=UStuttGrey, line width=0.2ex] (0,0) circle (0.5\satdiameter);
	}%
}

\title{\LARGE \bf
Tensegrity-based Robot Leg Design with Variable Stiffness}

\author{Erik Mortensen$^\dagger$, Jan Petrš$^\dagger$, Alexander Dittrich$^\dagger$ and Dario Floreano, \textit{IEEE Fellow}
\thanks{$\dagger$ These authors contributed equally to this work.}
\thanks{All authors are with the Laboratory of Intelligent Systems, École Polytechnique Fédérale de Lausanne (EPFL), Lausanne, Switzerland. Correspondance: {\tt\small \{erik.mortensen, jan.petrs, alexander.dittrich, dario.floreano\}@epfl.ch}}%
}

\usepackage{fancyhdr}
\begin{document}

\raggedbottom

\maketitle
\pagestyle{empty}

\thispagestyle{fancy}

\begin{abstract}

Animals can finely modulate their leg stiffness to interact with complex terrains and absorb sudden shocks. In feats like leaping and sprinting, animals demonstrate a sophisticated interplay of opposing muscle pairs that actively modulate joint stiffness, while tendons and ligaments act as biological springs storing and releasing energy. Although legged robots have achieved notable progress in robust locomotion, they still lack the refined adaptability inherent in animal motor control. Integrating mechanisms that allow active control of leg stiffness presents a pathway towards more resilient robotic systems.

This paper proposes a novel mechanical design to integrate compliancy into robot legs based on tensegrity - a structural principle that combines flexible cables and rigid elements to balance tension and compression. Tensegrity structures naturally allow for passive compliance, making them well-suited for absorbing impacts and adapting to diverse terrains. Our design features a robot leg with tensegrity joints and a mechanism to control the joint's rotational stiffness by modulating the tension of the cable actuation system. 
We demonstrate that the robot leg can reduce the impact forces of sudden shocks by at least \SI{34.7}{\percent} and achieve a similar leg flexion under a load difference of \SI{10.26}{\newton} by adjusting its stiffness configuration. The results indicate that tensegrity-based leg designs harbors potential towards more resilient and adaptable legged robots.

\end{abstract}

\section{Introduction}

Tensegrity robots have gained attention for their high strength-to-mass ratio, passive compliance, and structural versatility~\cite{shah2022tensegrity}. Tensegrity systems, composed of flexible tendons and rigid struts, naturally exhibit compliance, allowing them to absorb impacts and adapt to external forces~\cite{skelton2009tensegrity}. This inherent compliance makes tensegrity an attractive framework for building robots that are both lightweight and resilient \cite{gomez2020tensegrity,pugh1976introduction,kobayashi2022}. However, while tensegrity provides a natural ability to handle variability in the environment, there are tasks that demand more precise control over the robot’s stiffness, e.g. object manipulation. 
The combination of tensegrity's passive compliance with active, controllable stiffness mechanisms creates a powerful synergy that enables robots to adapt even more effectively to changing operational demands.

Prior research has highlighted the importance of variable stiffness in robotics for enhancing adaptability and robustness~\cite{kim2013soft,badri-sprowitz2022, sprowitz2013}. Systems that can actively tune their stiffness offer distinct advantages in managing dynamic interactions~\cite{laschi2016soft}. Tensegrity structures, with their capacity for structural compliance, are particularly well-suited for integrating variable stiffness mechanisms~\cite{zappetti2020variable}. By actively adjusting stiffness, robots can benefit from the best of both worlds: maintaining natural compliance when absorbing shocks or external forces, and increasing rigidity when precise control is needed \cite{shepherd2011multigait,sabelhaus2015system}.
\begin{figure}[t]
    \centering
    \includegraphics[width=1\linewidth]{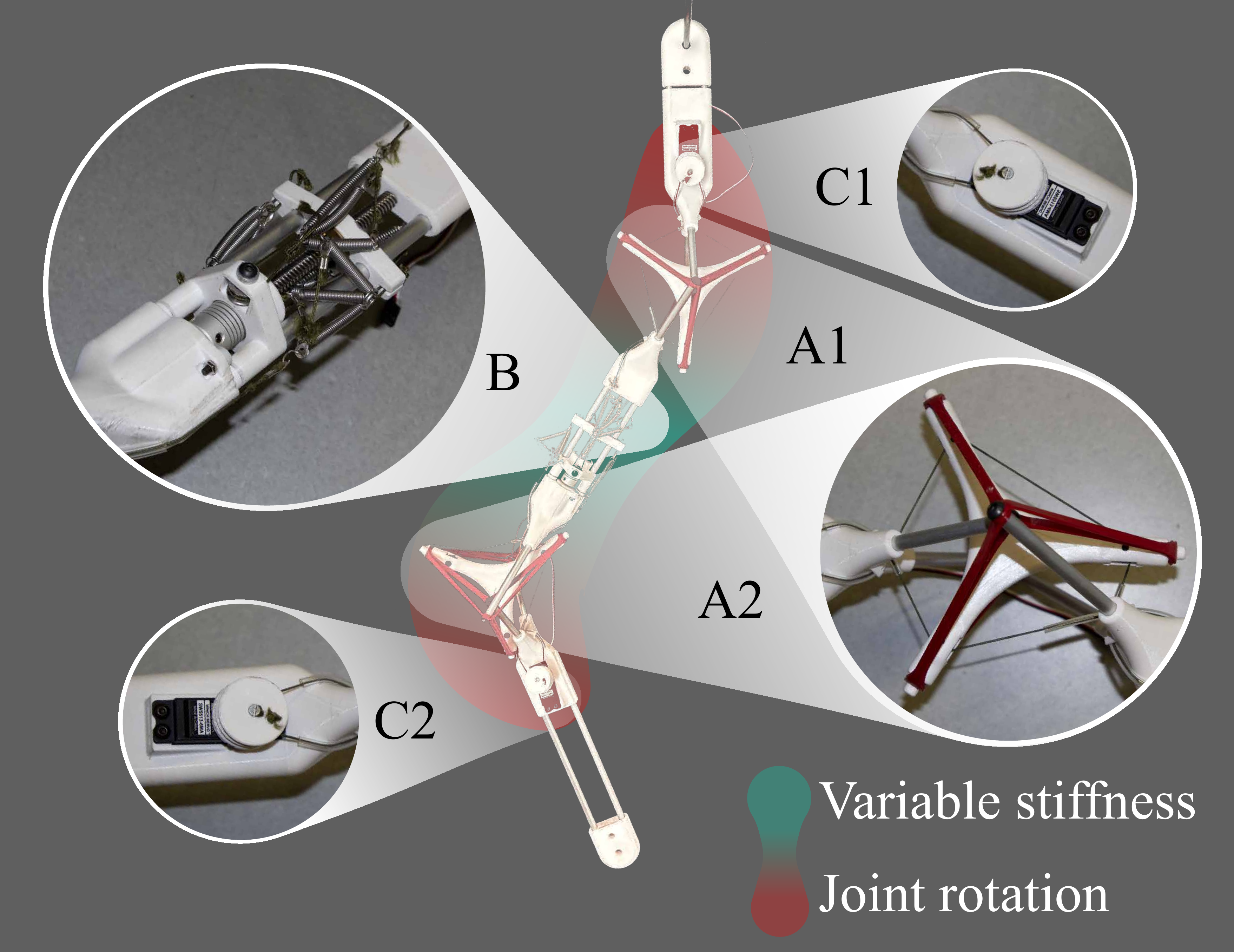}
    \caption{Robot leg design incorporating cable-driven actuation with a variable stiffness mechanism \Circled{A1}, \Circled{A2} tensegrity-based joints, \Circled{B} shared variable stiffness mechanism for both joints, \Circled{C1}, \Circled{C2} winch servos for controlling joint rotation.}
    \label{fig:robot-leg}
\end{figure}
Researchers investigated tensegrity-based joint designs~\cite{lessard2016b,friesen2018,jung2018,xie2022}, yet these approaches often feature limited range of motion (ROM), lack integrated variable stiffness, or involve complex kinematics and mechanical design that hinder their application in mobile robotics.
\begin{figure*}[!th]
    \centering
    \includegraphics[width=0.9\linewidth]{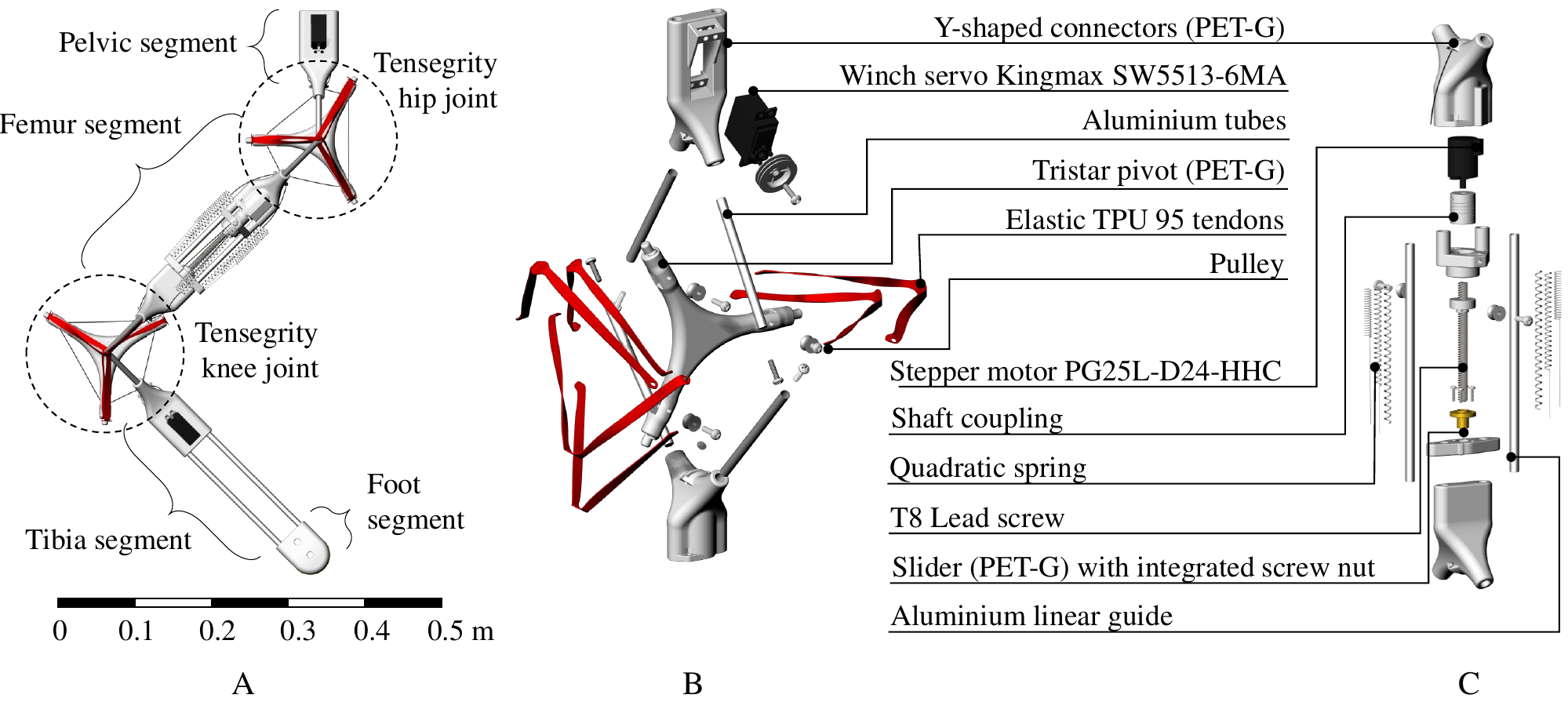}
    \caption{Proposed mechanical design of robot leg with \Circled{A} overview of main assembly with core sub-elements, \Circled{B} explosion graph of tensegrity joint with traversing winch actuation, \Circled{C} explosion graph of variable stiffness mechanism with lead screw and multi-spring system.}
    \label{fig:front-side-views}
\end{figure*}
This paper presents a robot leg design with two one-degree-of-freedom tensegrity joints and a mechanism to modulate the rotational joint stiffness (Figure~\ref{fig:robot-leg}). 
Traversing winch actuation drives the joint rotation. The variable stiffness mechanism modulates the joints' rotational stiffness by pre-tensioning the actuation cable system using a linear actuator. The tensegrity joints provide omnidirectional compliance and the variable stiffness mechanism can allow legged systems to dynamically adjust their response to external forces, enhancing shock absorption and providing further adaptability. 
Our design provides a large ROM of \SI{175}{\degree} with a single axis of rotation, simplifying controller synthesis. The actuation is fully integrated into the leg, enabling usability in mobile robot applications. The leg scale is comparable to a human leg with a total extended length of \SI{83}{\centi\meter}.

Our contributions in this paper are fourfold:
(i) a mechanical design for a cable-driven tensegrity-based rotary joint, 
(ii) a variable stiffness mechanism to control the rotational stiffness of the tensegrity-based joint,
(iii) the integration of these systems into a robotic leg,
(iv) an experimental characterization investigating the shock absorption capabilities and response to deformation of the proposed leg design with tensegrity joints and variable stiffness mechanism.

This work builds on advances in tensegrity robotics and variable stiffness, contributing a concept for legged robots that can flexibly adjust to dynamic environments through the symbiosis of passive compliance and active stiffness modulation.

\section{Related Works}

Our design builds on the work of Lessard et al.~\cite{lessard2016b}, who developed a bio-inspired tensegrity joint with passive compliance. However, their joint’s two-axis rotation poses challenges for robot control. By redesigning the pivot element from a T-shape to a tristar shape and introducing an overlapping V-expander configuration~\cite{muhao2023}, we achieve a single, cleaner rotation axis and an expanded motion range, enhancing its suitability for robotics.
Friesen et al.~\cite{friesen2018} proposed a tensegrity shoulder joint with variable stiffness through controller adjustments rather than a mechanical solution. While robust, its base-mounted actuation and weight make it less suitable for mobile platforms. Jung et al.~\cite{jung2018} explored a simple V-expander-based knee and hip joint inspired by human leg gait. Although Jung’s design lacks integrated actuation and variable stiffness, it demonstrates the feasibility of tensegrity in human-inspired movement.  
Zappetti et al.~\cite{zappetti2020variable} highlights the potential of variable stiffness with a tensegrity spine, which modulates stiffness by adding tendon tension. Their cable-driven mechanism provides discrete stiffness states through a ball joint, while our design extends this idea to continuous stiffness modulation without requiring contact-based mechanisms, enhancing adaptability.
Xie et al.~\cite{xie2022} employ a similar stiffening mechanism to ours by pre-tensioning cables with a screw drive. However, they focus on a tubercle-based joint with a limited ROM and high design complexity. Our joint, by contrast, is a single-axis design capable of large motion range and general-purpose adaptability, with a simple and compact mechanical integration.
In summary, prior designs have either complex  rotational control or lack active stiffness modulation. Our joint combines a \SI{175}{\degree} motion range with a defined rotation axis, and continuous stiffness modulation, advancing tensegrity’s applicability towards mobile robotic systems.

\section{Mechanical Design}

The proposed leg design is composed of two rotary tensegrity joints and three rigid segments. We refer to them as the pelvic segment, hip joint, femur segment, knee joint, and tibia segment. Figure~\ref{fig:front-side-views}~\Circled{A} provides an overview of the mechanical design. The variable stiffness mechanism is integrated inside the femur segment, modulating the stiffness of both tensegrity joints, while a traversing winch actuator controls the joint's angular position for each joint.

\subsection{Tensegrity Joint Design}
\label{sec:joint}

The tensegrity joint is composed of two rigid Y-shaped connectors, one rigid tristar pivot, and elastic tendons (Figure~\ref{fig:front-side-views}~\Circled{B}). The tendons connect the Y-connectors and the end of the tristar pivot via pins on the ends of the tristar pivot and a screw connection on the end of the Y-connectors. The tendons hold the tristar pivot in place, forming a tensegrity structure. The ends of the Y-connectors align with the center point of the tristar pivot, resulting in a single rotational axis for the joint.
For overlapping on the rotational plane, the Y-connectors have different branching angles, with the inner connectors branching angle having \SI{60}{\degree} and the outer \SI{80}{\degree}. The passive joint design without actuation directs rotational motion around the rotational axis with small counter torques introduced by tendon torsion. The tensegrity structure restricts and cushions motions that do not follow the joint's rotational axis. The tendon cross-section and the level of pre-tension in the tendons determine the omnidirectional dynamics of the joint and can be configured context-specific.

\begin{figure}[thb]
    \centering
    \includegraphics[width=0.7\linewidth]{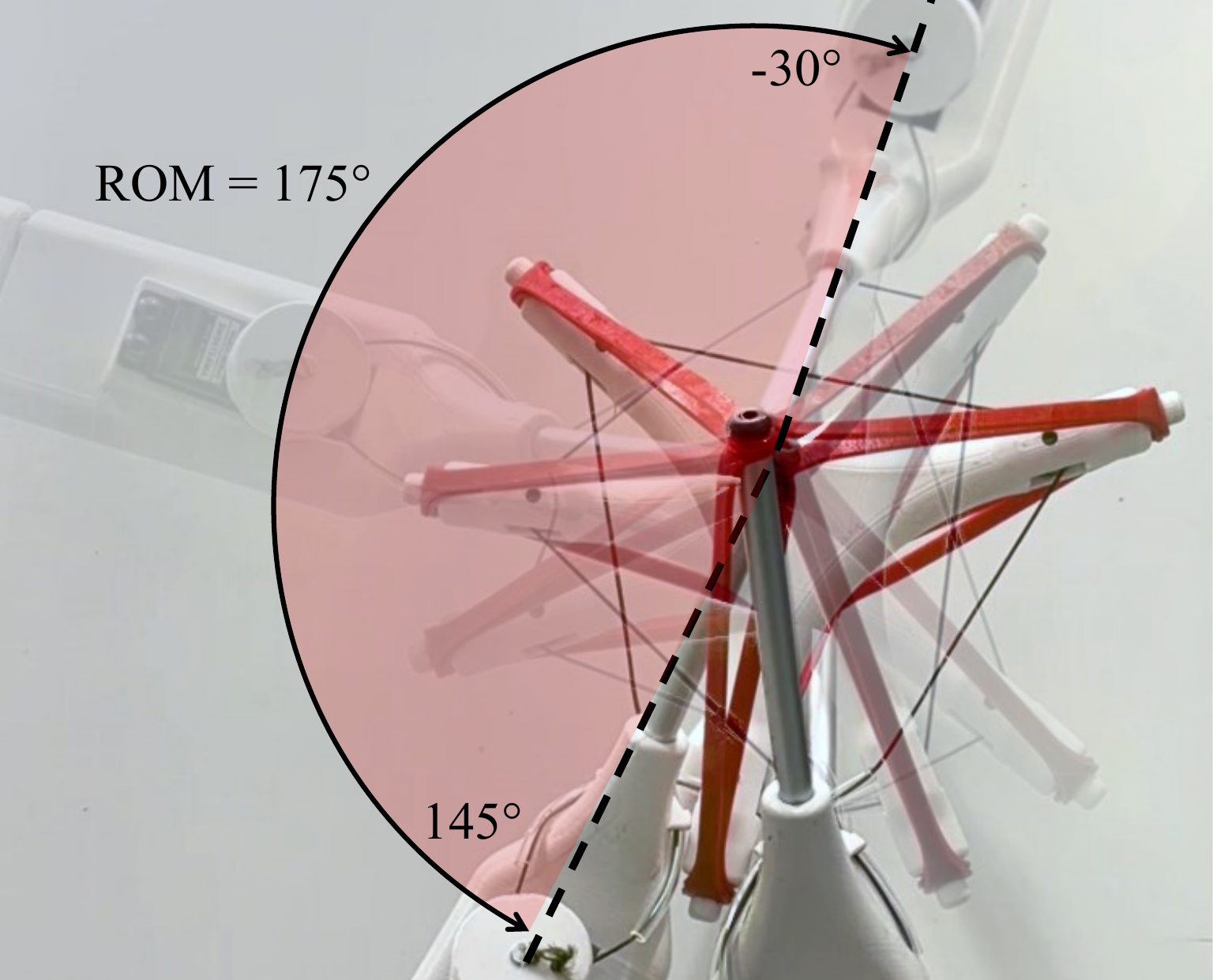}
    \caption{Rotation of tensegrity joint from minimum (-\SI{30}{\degree}) to maximum displacement (\SI{145}{\degree}).}
    \label{fig:joint-displacement-rom}
\end{figure}

In terms of fabrication, the Y-shaped connector and the tristar pivot are 3D printed with polyethylene terephthalate glycol (PET-G). Two aluminum tubes are inserted inside each connector to increase connector width and rigidity. The elastic tendons are 3D printed in  \textit{Thermoplastic Polyurethane 95a} (TPU 95). Zappetti et al.~\cite{zappetti2020variable,zappetti2017bio} show that this fabrication technique of the elastic cable network allows an automatized and high-fidelity fabrication of tensegrity structures. Each tendon incorporates three loops which interlock with one arm of the tristar pivot and each side of one of the connectors. The necessary tension to hold the tensegrity structure in place is obtained by pre-tensioning the tendons. We introduce pre-tension, by fabricating the tendon with reduced length in comparison to the designed geometric distance between the attachment points. We found that tendons with a cross-section of \SI{12}{\square\milli\meter} and a shrinkage factor of \SI{80}{\percent} provides a reasonable trade-off between sufficient elasticity to perform shock mitigation and sufficient rigidity to maintain structural integrity for joint functionality.

\subsection{Joint Actuation}
\label{sec:actuation}

Each joint is actuated through a system of cables and pulleys, connected to a winch servo motor. Figure~\ref{fig:joint-displacement-rom} shows the tensegrity joint and cable routing at different actuated positions. 
A double-sided winch is fixed on the motor, and attached to two antagonistic actuation cables, allowing it to pull on one when simultaneously releasing the other. These cables, made of high-strength nylon-polyester string (Atwood \SI{1.18}{\milli\meter} Micro Cord), are guided through the joint's tensegrity structure, using pulleys integrated inside the tristar pivot. These pulleys, located in the arms of the tristar pivot, enable the actuation of the joint. The actuation principle functions similarly to a cross belt drive. The other end of the actuation cables is fixed to the variable stiffness mechanism described in Section~\ref{sec:var-stiffness}.
Instead of the traversing winches also two translational actuators like linear electric actuators, pneumatic actuators e.g. McKibben actuators, or roller screw-based actuators could be utilized. Pneumatic actuation requires additional equipment for pressure generation. The winch mechanism requires only one instead of two actuators, simplifying control and mechanical integration and reducing costs.
The cable network of the traversing winches directs rotational motion in the joint, resulting in a one-degree-of-freedom rotational joint, assuming high-stiffness tendons and low displacement of the rotational axis. In practice, due to the compliance of the tendons of the tensegrity joint, omnidirectional motion within the joint is possible. These characteristics allow the mitigation of unseen perturbations.

For the prototype presented in this paper, we assumed a simplified static case of the leg holding a payload of $m = \SI{1}{\kilogram}$. The maximum load scenario for this case is  a fully extended lever arm of $l = \SI{0.83}{\meter}$ from the axis of rotation of the hip joint to the payload. 

To compute the required torque for the joint actuation, we compute the moment in the hip joint and add the reduction ratio by the winch mechanism. The cable actuation mechanism can be approximated as two gears with a reduction ratio of 1:12, based on the winch drum radius of $r_{\text{drum}} = \SI{5}{\milli\meter}$ and the radius of deflection pulleys in the tristar pivot  $r_\text{tristar} = \SI{60}{\milli\meter}$ towards the axis of rotation of the joint. 


The required torque is

\begin{equation}
\tau_{\text{req}} = \frac{1}{2} m g l \cdot k_s \cdot \frac{r_\text{drum}}{r_\text{tristar}} = \SI{0.664}{\newton\meter}
\end{equation}

where $g$ is the gravitational constant. 

We selected the off-the-shelve, small form factor, low-cost winch servo Kingmax SW5513-6MA. As specified by the manufacturer, it provides a motion range of up to \SI{2160}{\degree} with a torque of \SI{1.042}{\newton\meter} to suffice for the calculated required torque $\tau_{\text{req}}$. 


\subsection{Variable Stiffness Mechanism}
\label{sec:var-stiffness}

The variable stiffness mechanism allows active control of the joint stiffness of the robot leg by changing the tension in the cables. To adjust the joint stiffness, we connect a multi-spring system in series with each actuation cable of the traversing winch actuation in the antagonistic configuration. As depicted in Figure~\ref{fig:front-side-views}~\Circled{C}, both cables are mounted at a slider with an integrated screw nut of a lead screw (S~SIENOC~\SI{100}{\milli\meter}~T8). By changing the position of the slider with a stepper motor (Minebea-Mitsumi PG25L-D24-HHC1), we can modify the tension in the actuation cable and spring, which influences the rotational stiffness at the joint. For mechanical packaging, we integrated one lead screw mechanism in the femur of the robotic leg, controlling the stiffness of both, the hip and knee joint. The second joint actuation is attached via an additional pulley, equally adding tension to both joints via one translation motion of the lead screw. While the stiffness of each joint could be controlled individually by integrating two variable stiffness mechanisms, we decided to control the stiffness of both joints with one variable stiffness mechanism. This choice results in a more compact design and lower weight in the overall leg. Figure~\ref{fig:front-side-views} shows an overall view of the elements composing the variable stiffness mechanism.

\begin{figure}[tb]
    \centering
    \includegraphics[width=0.65\linewidth]{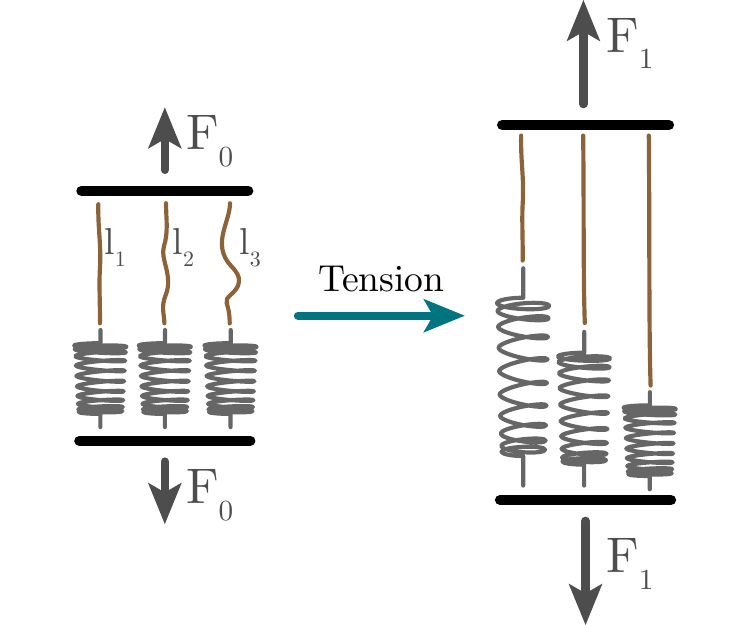}
    \caption{Working principle of the multi-spring system to achieve quadratic stiffness behavior. Tension springs are in series with cables of different lengths, adding backlash in the cables.}
    \label{fig:spring-system}
\end{figure}

\begin{figure}[t]
    \centering
    \includegraphics[width=0.85\linewidth]{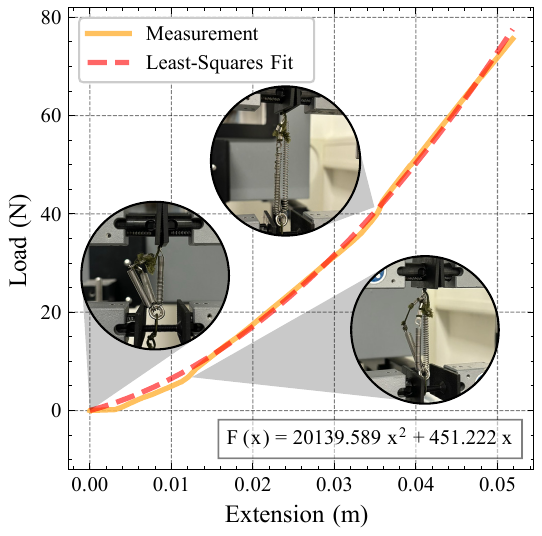}
    \caption{Fitting of unbiased quadratic function as approximation of the multi-spring system.}
    \label{fig:quadratic_spring}
\end{figure}

We design a multi-spring system approximating quadratic spring behavior by using three tensile springs in parallel.
English and Russel~\cite{english1999} analyzed antagonistic variable stiffness actuation. They conclude that decoupled joint stiffness from joint deflection requires quadratic load to extension behavior. Otherwise, the tension forces generated by the springs would cancel each other. The selected stainless steel springs have a length \SI{35}{\milli\meter}, a diameter of \SI{4.9}{\milli\meter} and wire thickness of \SI{0.7}{\milli\meter}, providing a linear stiffness of approximately \SI{388.4}{\newton\per\meter}. 
Each spring is attached to the nylon-polyester cable in series with different lengths.
Using this length difference, the three springs engage at different stages of elongation. When the tension is low, only one spring is engaged. The second spring engages when the elongation of the first spring exceeds the length of \SI{12}{\milli\meter} and the third spring when the systems' elongation exceeds \SI{36}{\milli\meter}. Figure~\ref{fig:spring-system} illustrates this principle. 
To confirm the multi-spring systems' quadratic behavior, we measure the load-to-extension ratio. Figure~\ref{fig:quadratic_spring} shows the measurement compared to a quadratic fit of an unbiased quadratic function using a non-linear least square curve fit. 
Within the allowed extension of one individual spring, the function shows a nearly perfect quadratic behavior, with a coefficient of determination of $R^2 = 0.9983$.

\section{Experimental Characterization}

This paper assesses the variable stiffness capabilities of a prototype of the proposed robot leg design. We propose two experiments: In the first experiment in Section~\ref{sec:variable-stiffness}, we statically measure the legs' stiffness in a tensile testing machine, to assess the influence of the variable stiffness mechanism under applied external load. 
The second experiment in Section~\ref{sec:drop-experiment} investigates the dynamic response of the leg to a sudden impact in a drop experiment. 
For both experiments, we examine three different stiffness settings by configuring the displacement of the slider of the variable stiffness mechanism. In the minimum stiffness setting, we have \SI{0}{\milli\meter} initial displacement in the multi-spring system. The medium setting moves the initial attachment point to \SI{20}{\milli\meter}, while the maximum stiffness configuration sets \SI{40}{\milli\meter} linear displacement. The initial joint flexion in both experiments is \SI{+25}{\degree} in the hip joint and \SI{-25}{\degree} in the knee joint from a fully stretched alignment.

\begin{figure*}[t]
    \centering
    \includegraphics[width=.85\linewidth]{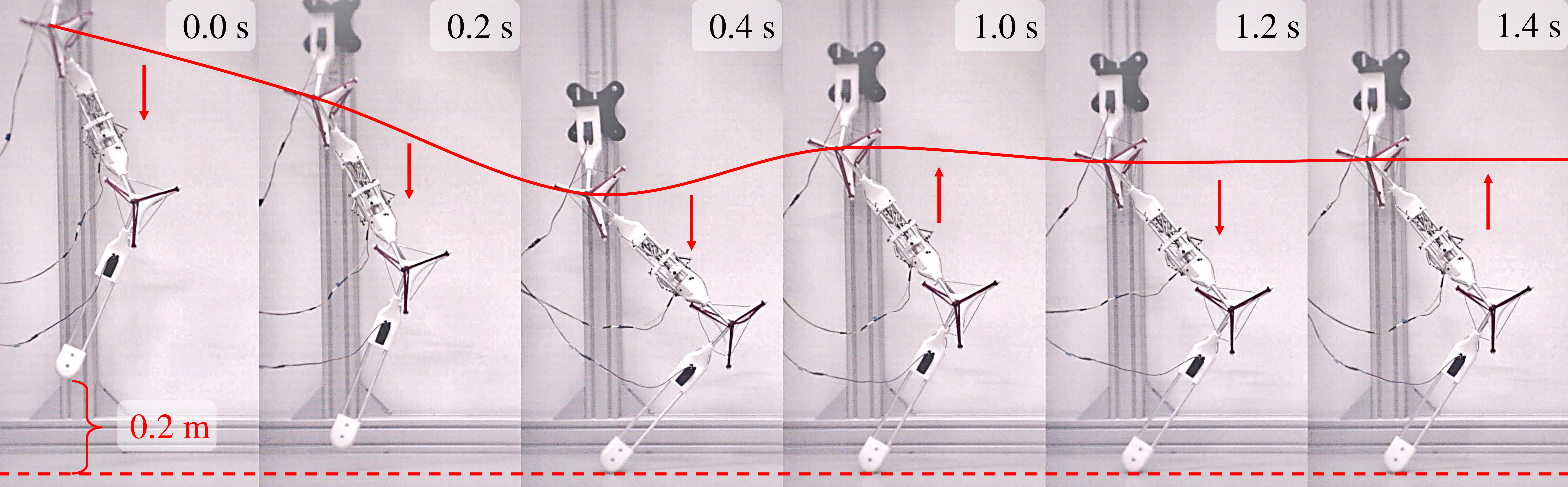}
    \caption{Times series sequence of the drop experiment showing the robot leg fixed at the linear guide during the dropping motion. The red line qualitatively indicates the position change of the hip joint's axis of rotation rigidly connected to the tracked pelvis segment. After ground contact, a notable flexion of hip and knee joint can be observed before the leg reaches steady-state position. Arrows signal the motion direction of the fixed pelvis segment.}
    \label{fig:drop_sequence}
\end{figure*}

\subsection{Leg deformation under compression} 
\label{sec:variable-stiffness}

\begin{figure}[ht]
    \centering
    \includegraphics[width=0.95\linewidth]{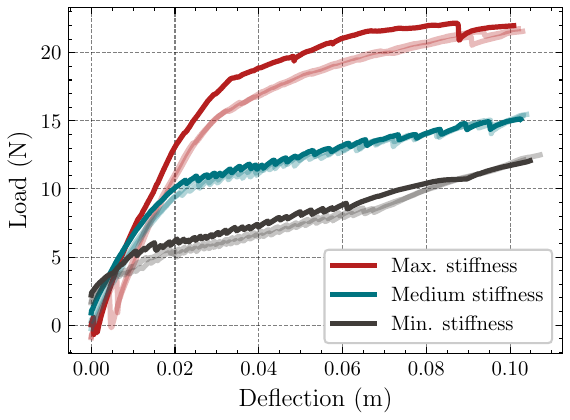}
    \caption{Measurement of reaction force for minimum, medium, and maximum stiffness level, during compression test. For each level, three extensions to load curves were measured. For better readability, two measurements have increased transparency.}
    \label{fig:variable-stiffness}
\end{figure}

We use the Instron~5960 series tensile machine. We fixate the robot leg at the pelvis segment in the clamping fixture at the load cell of the testing machine. At the start of the experiment, the foot has contact with a ground plate without any compression forces applied. A rubberized anti-slip pad is attached to the foot segment's tip to prevent slip during the experiment. The tensile machine vertically compresses the robot leg by approximately \SI{10}{\centi\meter} with a low speed of \SI{0.5}{\milli\meter\per\second} and measures forces exerted from the robot leg to the load cell. We perform this experiment three times for each stiffness setting. Figure~\ref{fig:variable-stiffness} shows the measured deflection to load curves.

One can observe, three distinct load-to-deflection characteristics for each stiffness setting. For each setting, the exerted forces on the load cell are increasing, showing an increase in the rotational stiffness of the joints. At \SI{0.1}{\meter} compression, the leg exerts a reaction force of \SI{11.71}{\newton} in the minimum, \SI{15.00}{\newton} in the medium and \SI{21.97}{\newton} in the maximum stiffness setting. 
This difference of \SI{87.61}{\percent} implies a higher capacity to apply forces in the maximum stiffness setting, for example in a heavy lifting task.
Furthermore, oscillatory behaviors can be observed, partially due to non-uniform tensioning of the antagonistic and variable stiffness cables. This inconsistency can be attributed to localized friction between the cables and other structural components of the leg and torsion in the tendons. For example, improvements in cable routing and tendon attachment could reduce these oscillations.

\subsection{Shock mitigation after drop}
\label{sec:drop-experiment}

\begin{figure}[tbh]
    \centering
    \includegraphics[width=0.89\linewidth]{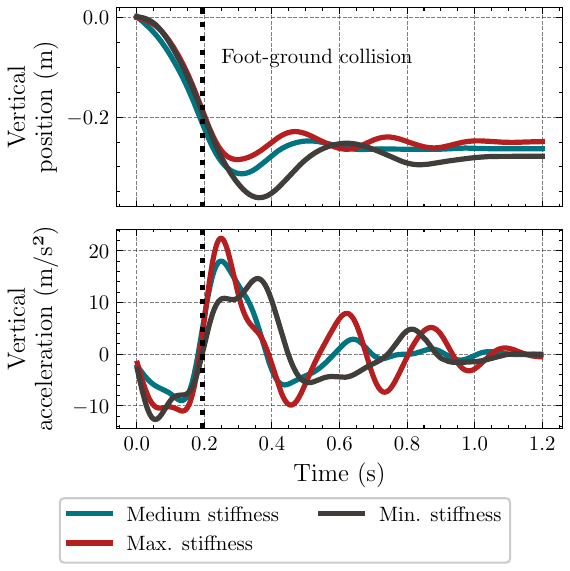}
    \caption{Filtered change of vertical position of the pelvic segment during the drop experiment and derivated vertical acceleration for each stiffness setting.}
    \label{fig:drop-experiment}
\end{figure}

We evaluate the shock mitigation capabilities in a series of drop tests. We attach the robot leg at a linear guide with a cart, resulting in an additional mass of \SI{254}{\gram} at the pelvis segment. 
The joint actuation of both joints applies a constant holding torque during the drop experiment without further control. We drop the leg with a distance of \SI{20}{\centi\meter} between the foot tip and a solid ground plate. Figure~\ref{fig:drop_sequence} shows a time sequence of the drop experiments. The high-speed camera Chronos CR21-1.0-16C records the drop with \SI{1000}{\hertz}. We extract the vertical position of the pelvic segment by using an open-source video tracking software \cite{brown2008video} and apply central finite elements to obtain the vertical acceleration. A Savitzky-Golay and a Butterworth filter remove measurement noise from both signals. 
Figure~\ref{fig:drop-experiment} shows the vertical position and vertical acceleration over time for the different stiffness settings.
For all three stiffness settings, the foot remained in contact with the ground after collision, showing shock cushioning.
The vertical position of the pelvis segments shows a higher deflection after the collision in the minimum stiffness with a deflection of \SI{-0.278}{\meter} in negative vertical direction compared to \SI{-0.263}{\meter} in the medium and \SI{-0.249}{\meter} in the maximum stiffness setting. It can be assumed, that the change in the joints' rotational stiffness due to the change of pre-tension in the multi-spring system is the cause for this change in behavior. This change in stiffness also causes a higher steady-state deflection for settings with lower stiffness setting, with a deflection of \SI{-0.277}{\meter} compared to \SI{-0.249}{\meter} for the stiffest setting. Dynamically, one can also observe a longer deceleration phase for the minimum stiffness setting, with a duration of \SI{0.170}{\second} compared to \SI{0.101}{\second} at maximum stiffness. 
Meanwhile, the maximum stiffness setting results in a more periodic, damped oscillatory motion with larger acceleration amplitudes and frequency. 
The change of rotational stiffness is particularly notable for the magnitude of vertical acceleration. The maximum stiffness setting results in a peak vertical acceleration of \SI{22.43}{\meter\per\square\second}, while the minimum stiffness setting shows a peak acceleration of \SI{14.65}{\meter\per\square\second}. This difference is a significant reduction of \SI{34.7}{\percent}, which correlates linearly to lower impact forces at the pelvic segment after collision.
The minimum stiffness configuration increases deformation and reduces peak acceleration, which contributes to improving the system's stability and reducing damage from sudden impacts. The maximum stiffness configuration, on the contrary, decreases deformation, which enhances higher precision in position control and better exertion of large forces, such as handling high payloads or intentionally impactful environment interactions.

\section{Conclusion}

This paper presents a robotic leg design with tensegrity joints and an integrated variable stiffness mechanism.



Through the compression experiment, we demonstrate that a \SI{10}{\milli\meter} vertical deformation produces a reaction force difference of up to \SI{10.26}{\newton} between the minimum and maximum stiffness settings, showing the variable stiffness mechanism's ability to control deformation under external forces. This adaptability allows for softer behavior when flexibility is beneficial and increased stiffness in conditions requiring resistance to high compression forces, such as payloads.
In a dynamic drop experiment, we observed that adjustments in the stiffness configuration can reduce peak accelerations of at least \SI{34.7}{\percent}. This reduction in peak acceleration, translating to a decrease in impact forces and larger deformation, enhances damage mitigation during sudden impacts.




However, the current design's application to real-time locomotion requires further improvement towards faster actuation integration and reducing counter torques by friction and tendon torsion. To mitigate these issues, future designs can consider high-performance brushless motors or dynamic actuators like McKibben actuators. Additionally, precise control is constrained by the lack of feedback sensors; integrating joint encoders, load cells, and acceleration sensors would enable more refined control over both movement and stiffness adjustments. 

Despite these limitations, the prototype demonstrates the potential of variable stiffness for robotic legged systems. Further design iterations will in all likelihood facilitate even larger differences in stiffness modulation and shock mitigation beyond the results demonstrated with this paper’s initial prototype. Further work will be required to validate this concept in a full-scale multi-legged robot with sensor integration, enhanced actuation, and holistic control approaches; leading to demonstrate a robot that shows the prospect of tensegrity joints combined with variable stiffness for robot locomotion in complex and uncertain environments.

\section{Acknowledgements}
This work was supported by the Swiss National Science Foundation (SNSF) and the Japan Society for the Promotion of Science (JSPS) under project number \texttt{IZLJZ2\_214053}. 






\bibliographystyle{ieeetr}
\bibliography{references}

\newpage
\appendices
\section{Joint Characterization}

To evaluate key mechanical properties of the tensegrity joint and determine an beneficial tendon configuration, two experimental tests were conducted. These tests characterize the joint’s \emph{passive resistance to rotation} and \emph{passive resistance to coaxial compression}. We conduct this experiments without actuation system attached, so that the measured resistance arises solely from tendon deformation of the tensegrity joint.

Four tendon configurations were tested, varying in cross-sectional area (\SI{8}{\milli\meter\squared} and \SI{12}{\milli\meter\squared}) and pre-tension levels (70\%, 80\%, and 90\% relative to the unstretched tendon length in the tensegrity structure). However, the \SI{70}{\percent} \SI{12}{\milli\meter\squared} configuration was excluded due to excessive tightness, and the \SI{90}{\percent} \SI{8}{\milli\meter\squared} configuration was omitted as it was too loose to maintain structural integrity. The tendon set minimizing rotational resistance while maximizing coaxial loading resistance was selected for subsequent experiments.

\subsection{Passive Joint Resistance to Rotation}

\begin{figure}[!h]
\centering
\subfloat[Photograph of the test]{\includegraphics[width=.45\columnwidth]{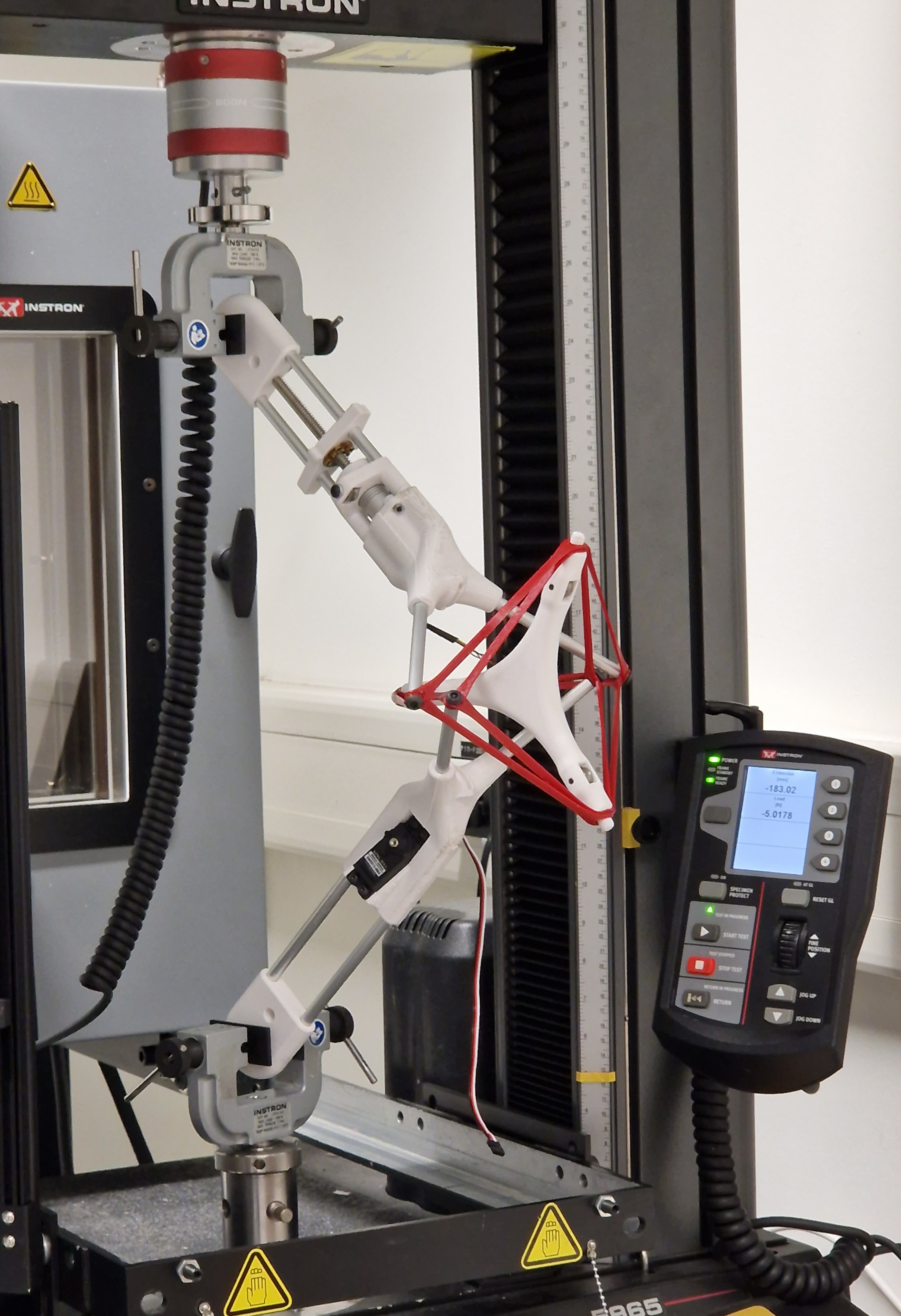}} \quad
\subfloat[Schematic representation]{\includegraphics[width=.45\columnwidth]{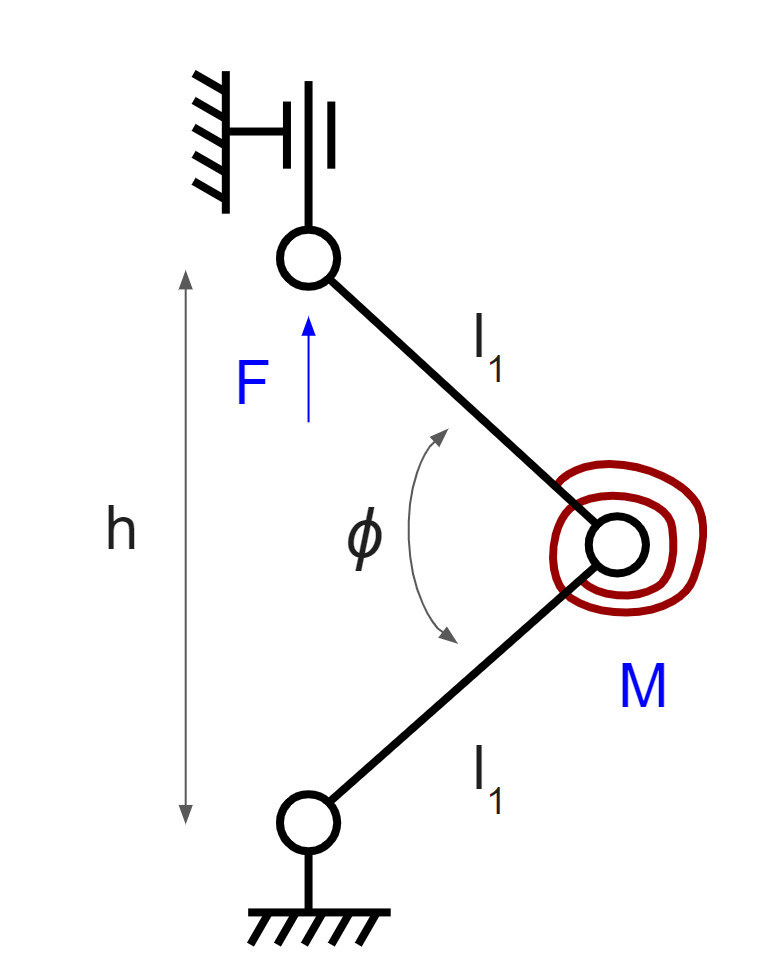}} \quad
\caption{Experimental setup for measuring the joint’s passive resistance to rotation.}
\label{fig:test_passive_rotation}
\end{figure}

To assess rotational resistance, the joint was mounted between two arms (each \SI{310}{\milli\meter} in length) and attached to a tensile testing machine, as shown in Figure \ref{fig:test_passive_rotation}. The joint’s rotation angle, $\phi$, was linked to the vertical displacement, $h$, by:

\begin{equation}
\phi = 2 \arcsin\left(\frac{h}{2 l_1}\right)
\label{eq:phi}
\end{equation}

The joint started at its equilibrium position ($\phi = \SI{120}{\degree}$) and was incrementally bent to a minimum of \SI{10}{\degree} at a rate of \SI{4}{\milli\meter\per\second}. The reaction force, $F$, measured by the tensile testing machine, was adjusted to reduce gravitational effects. Given that the joint behaves as a rotational spring, the torque required to rotate it was computed as:

\begin{equation}
M = F l_1 \cos\left(\frac{\phi}{2}\right)
\label{eq:torque}
\end{equation}

\begin{figure}[!h]
\centering
\includegraphics[width=.99\columnwidth]{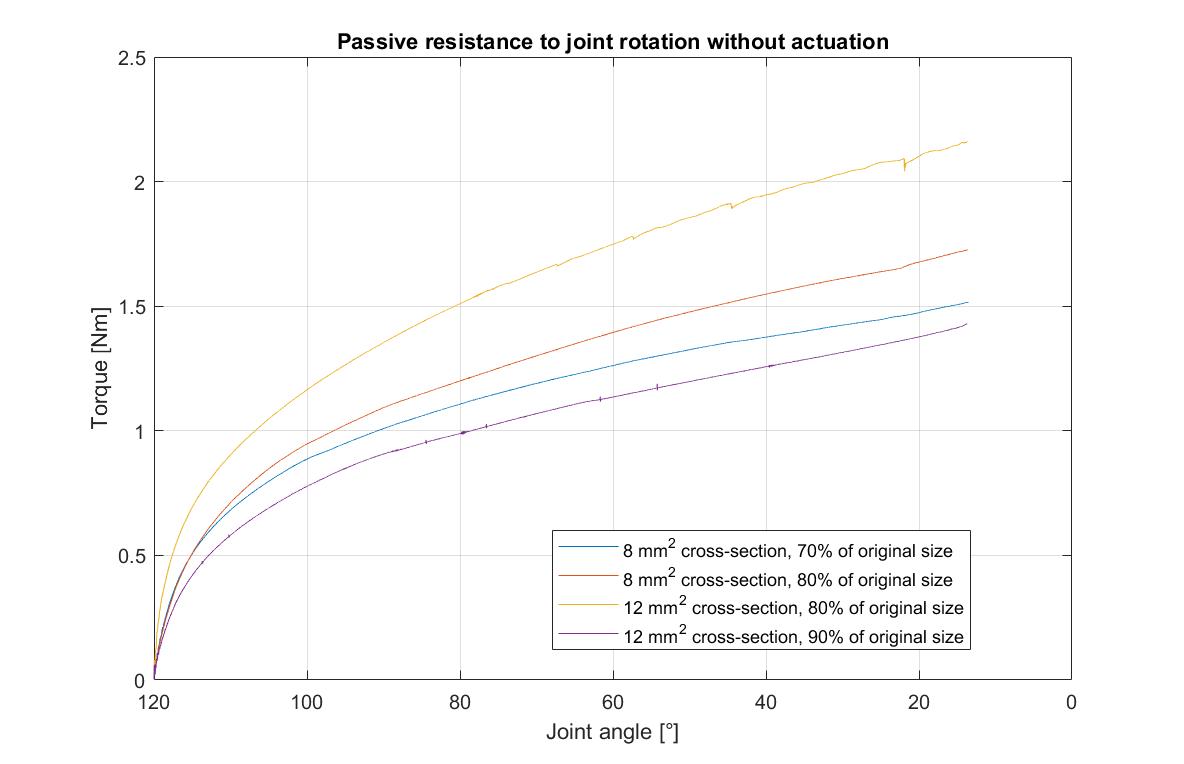}
\caption{Measured passive resistance to rotation for different tendon configurations.}
\label{fig:test_passive_resistance}
\end{figure}

The results, plotted in Figure \ref{fig:test_passive_resistance}, indicate that the joint exhibits increasing rotational resistance as it moves away from equilibrium. Additionally, increasing both tendon pre-tension and cross-sectional area raises passive rotational resistance.

The data suggest that a \SI{12}{\milli\meter\squared} cross-section with \SI{90}{\percent} pre-tension achieves the lowest rotational resistance while maintaining adequate joint stability.

\subsection{Passive Joint Resistance to Coaxial Loading}

\begin{figure}[!h]
\centering
\subfloat[Photograph of the test]{\includegraphics[width=.45\columnwidth]{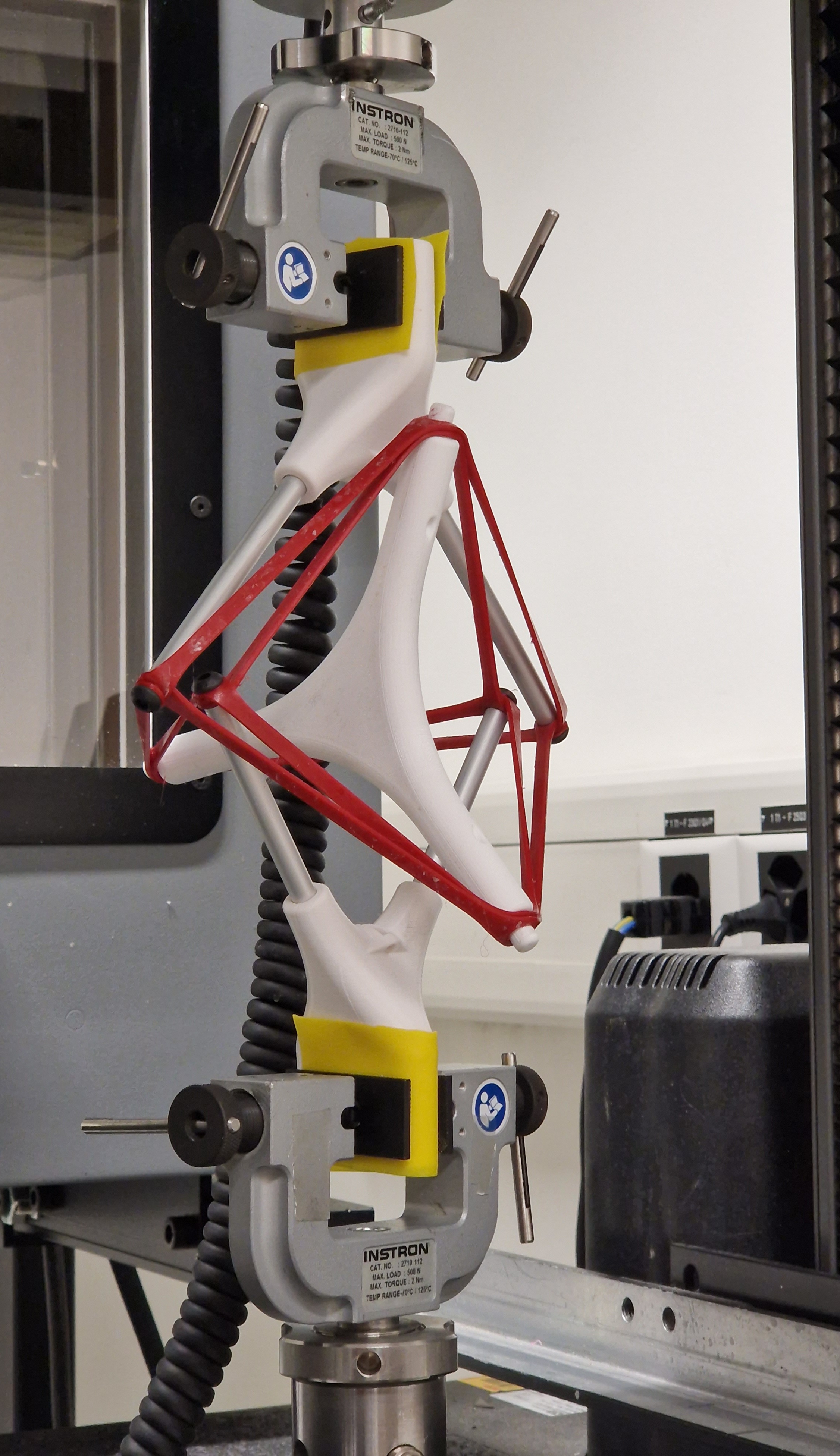}} \quad
\subfloat[Schematic representation]{\includegraphics[width=.45\columnwidth]{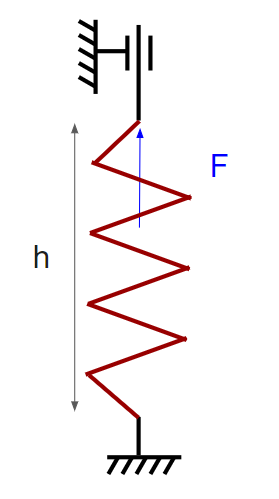}} \quad
\caption{Experimental setup for measuring passive resistance to coaxial compression.}
\label{fig:test_passive_compression_pictures}
\end{figure}

\begin{figure}[!h]
\centering
\includegraphics[width=.99\columnwidth]{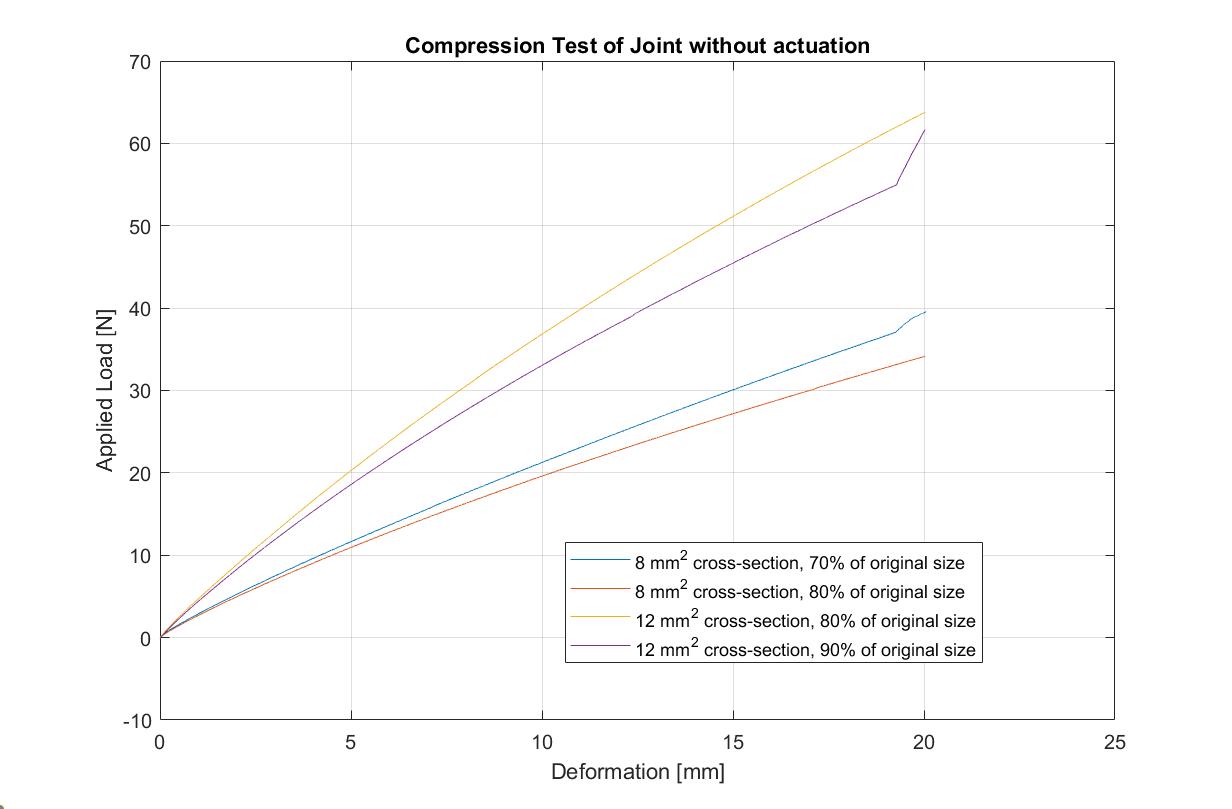}
\caption{Measured passive resistance to coaxial compression for different tendon configurations.}
\label{fig:test_passive_compression}
\end{figure}

To evaluate coaxial compression resistance, the joint was clamped between two fixed supports at a \SI{180}{\degree} angle (Figure \ref{fig:test_passive_compression_pictures}). The tensile testing machine compressed the joint at a constant rate of \SI{0.5}{\milli\meter\per\second} until rigid structural components contacted each other. Reaction forces were recorded for each tendon configuration, as shown in Figure \ref{fig:test_passive_compression}.

The results indicate that both increased cross-sectional area and higher pre-tension improve resistance to coaxial loading. The resistance appears to increase linearly with deformation. Larger cross-sectional areas (\SI{12}{\milli\meter\squared}) generally provide greater stiffness and higher resistance to compression.





\end{document}